\documentclass{article}

\PassOptionsToPackage{table}{xcolor}  

\usepackage[final, main]{neurips_2026}

\usepackage[utf8]{inputenc}
\usepackage[T1]{fontenc}
\usepackage{hyperref}
\usepackage{makecell}
\usepackage{url}
\usepackage{booktabs}
\usepackage{amsfonts}
\usepackage{amssymb}
\usepackage{amsmath}
\usepackage{nicefrac}
\usepackage{microtype}
\usepackage{xcolor}
 \usepackage{natbib}
\usepackage{graphicx}
\usepackage{subcaption}
\usepackage{multirow}
\usepackage{makecell}
\usepackage{enumitem}
\usepackage{latexsym}
\usepackage{marvosym}
\usepackage[normalem]{ulem}
\usepackage{hyperref}
\usepackage{url}
\usepackage{graphicx}
\usepackage{booktabs}
\usepackage{multirow} 
\usepackage{makecell}
\usepackage{wrapfig}
\usepackage{subcaption}
\usepackage{tabularx}
\usepackage{colortbl}
\usepackage[ruled,vlined]{algorithm2e}

\definecolor{posgreen}{RGB}{0, 153, 51}
\definecolor{negred}{RGB}{204, 0, 0}

\renewcommand{\arraystretch}{.9}

\title{Neuron-Aware Data Selection for Annotation-Free LLM Self-Distillation}

\author{%
  Zhuowei Chen  \hspace{1em} Xiang Lorraine Li \\
  University of Pittsburgh, United States \\
  \texttt{\{zhuowei.chen, xianglli\}@pitt.edu}
}

\begin{document}

\maketitle

\begin{abstract}

Post-training large language models (LLMs) without real-world interaction feedback or human-labeled supervision remains challenging, particularly in specialized domains where expert annotations are costly to obtain. Recent annotation-free self-evolution methods address this by using the model's own outputs as supervision signals, constructing a teacher via additional context and aggregating predictions across multiple rollouts through majority voting to produce pseudo-labels. However, these approaches are not without drawbacks: SFT- and GRPO-based variants suffer out-of-domain performance degradation, while reward-based on-policy RL inflates calibration error.
In this paper, we propose \textsc{Neuron On-Policy Self-Distillation (Neuron-OPSD)}, a data-centric framework for annotation-free self-distillation that leverages internal neuron activations to guide both training-data selection and teacher context construction. The model is then trained via on-policy distillation from the teacher distribution, requiring no ground-truth labels at any stage. Across specialized-domain benchmarks, \textsc{Neuron-OPSD} improves in-domain task performance while preserving cross-domain generalization and mitigating calibration collapse over prior annotation-free baselines. This framework is particularly relevant to settings where online interaction or external supervision is costly or infeasible, and is conceptually distinct from offline RL approaches that rely on logged, reward-labeled trajectories.
\end{abstract}

\begin{figure}[h]
    \centering
    \includegraphics[width=.9\linewidth]{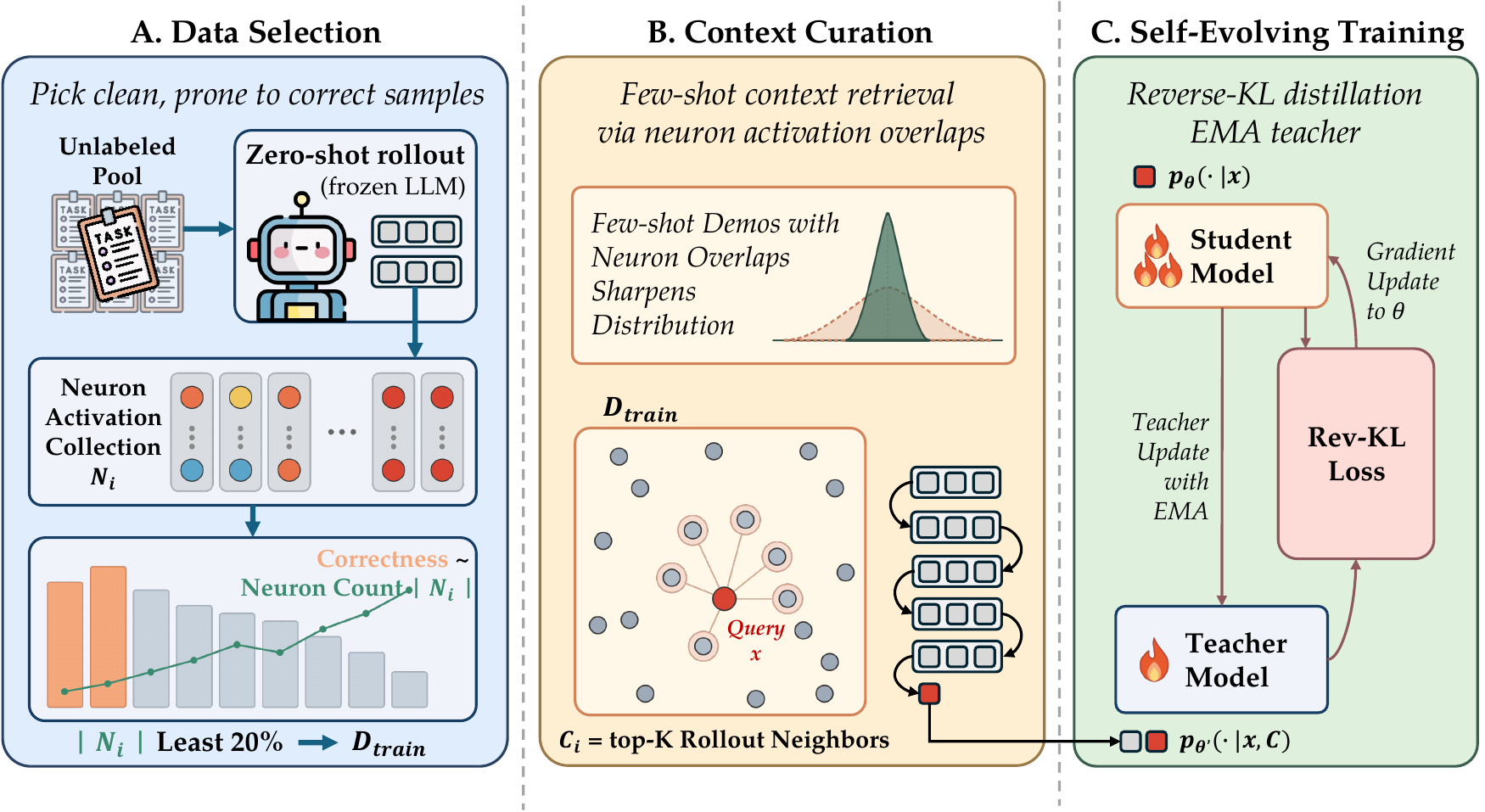}
    \caption{Overview of the Proposed Neuron-OPSD.}
    \label{fig:overview}
\end{figure}

\section{Introduction}

Recent advances in Large language models (LLMs) have demonstrated remarkable performances in general tasks, however, adapting them to highly specialized domains such as education, law, or STEM subjects, remains difficult, which mainly due to the scarcity of human expert annotations.

To mitigate the scarcity of expert annotations, recent work has explored {annotation-free self-training}, where the model improves itself using only unlabeled data and signals derived from the model itself. 
SFT-based methods such as LMSI fine-tune the model on its own generated responses or rationales~\citep{huang-etal-2023-large}. 
GRPO-based methods instead construct scalar rewards from the model's own outputs. TTRL extracts pseudo-labels from majority voting over multiple rollouts and optimizes them with the GRPO algorithm~\citep{zuo_ttrl_2025}, while Intuitor uses entropy-based confidence signals as rewards~\citep{zhao_learning_2025}. 
Despite their empirical successes, these annotation-free paradigms still face fundamental limitations. SFT-based self-training can suffer from catastrophic forgetting, especially for out-of domain tasks. GRPO-based optimization faces challenges such as vanishing gradients under low within-group reward variance and entropy collapse during prolonged training. More fundamentally, GRPO collapses each rollout into a scalar reward, yielding only trajectory-level supervision and offering no guidance on which part of the reasoning should be reinforced or revised.

On-policy distillation (OPD) offers finer supervision by training the student model against a teacher model's token-level predictive distribution rather than a hard final-answer reward. Such logit-level targets densify supervision along the generation trajectory and circumvent the learning collapse caused by low within-group reward variances~\citep{song2026surveyonpolicydistillationlarge}. However, OPD's effectiveness relies on the quality of the teacher distribution, which is typically derived from stronger models, verifiers, or curated demonstrations. 
When such external signals are unavailable, the teacher must be elicited from the model itself through additional information in prompts, i.e few-shot examples.
This reframes the problem from designing self-generated rewards alone to deciding which unlabeled samples to train on and which contexts induce a useful teacher–student gap.
Crucially, this self-improvement process requires rigorous data selection. Because the model acts as its own teacher, indiscriminately distilling from self-generated signals risks reinforcing its existing miscalibration, spurious reasoning, and hallucinations. 

While traditional selection methods rely on surface-level metrics like majority-voting or entropy to filter out such unreliable data, these signals often prove insufficient. 
Recent work has pivoted towards LLMs' internal dynamics, demonstrating that internal states, particularly neuron activations, exhibit a strong correlation with when a model is genuinely uncertain or hallucinating~\citep{gao2024scaling,nostalgebraist2020interpreting,chen2025llms,chen-etal-2026-neuron}.


Motivated by these observations, we propose \textsc{Neuron-Opsd} (N-OPSD), an annotation-free OPD pipeline for LLM self-improvement. In this work, we focus on a two-fold data construction process that determines whether self-distillation provides a useful signal: which unlabeled samples should be used for training, and which few-shot contexts should be used to induce the teacher distribution.

As illustrated in Figure~\ref{fig:overview}, \textsc{Neuron-Opsd} utilizes two neuron-based mechanisms. First, {Neuron Consensus} analyzes the reliability of candidate self-training samples through activation statistics. We measure neuron consensus using the number of activated neurons, finding that this signal is strongly associated with hallucinated or unreliable behaviors of LLMs. However, our results show that consensus alone is not a complete selection criterion. 
Selecting samples solely by activation count does not consistently yield stronger self-distillation, suggesting that training utility depends on both the reliability of the self-generated signal, and the room for teacher-induced sharpening, i.e., how much information can student model learn from the teacher. This suggests that useful training data must be reliable enough to avoid reinforcing hallucinations, yet uncertain enough to leave room for LLMs improvement.
Second, \textsc{Neuron-Opsd} employs {Neuron Overlap} for the OPD teacher-context construction. 
For each query, we retrieve the examples that show the most similar reasoning pattern to the query as demonstrations, which are measured using active neuron representations.
Since the teacher is the same model conditioned on an augmented context, these retrieved demonstrations directly shape the token-level distribution used for distillation. Neuron-overlap retrieval, therefore serves as a label-free method to elicit a more informative teacher distribution and create a meaningful teacher-student gap.
Together, these two components demonstrate how to effectively leverage internal signals from LLMs for annotation-free OPD. {Neuron Consensus} serves primarily as a diagnostic signal to identify which samples are reliable or improvable, while {Neuron Overlap} provides a practical mechanism for constructing robust teacher contexts. 
This makes \textsc{Neuron-Opsd} a practical data-centric pipeline focused strictly on selecting the optimal data and contexts to form high-quality supervision.

We conduct experiments across three datasets covering six source domains. Results show that \textsc{Neuron-Opsd} achieves competitive self-improvement while better preserving cross-domain generalization and calibration compared with previous baselines. 
Further analysis suggests that its gains are driven by both the LLM's initial capability on the training dataset and the teacher-student distribution gap induced by the contexts. Ablation studies on four SciKnowEval domains further show that neuron consensus data selection is informative but insufficient as a standalone rule, while neuron-overlap retrieval provides an effective context-curation mechanism.

Our core contributions are as follows:
\begin{itemize}[leftmargin=*, itemsep=0pt, topsep=0pt]
    \item We utilize {Neuron Consensus} as a label-free internal signal for hallucination detection and data selection. Activation counts correlate with hallucinated behavior, but selecting neither low- nor high-activation data consistently gives the best self-distillation results. This shows that activation count is informative, but insufficient as the only data-selection rule.

    \item We propose {Neuron Overlap} for OPD context curation. It retrieves few-shot demonstrations with similar active-neuron patterns, inducing a more informative teacher distribution and a useful teacher-student gap for annotation-free OPD.

    \item We integrate these two signals into \textsc{Neuron-Opsd}. Experiments across specialized-domain benchmarks show that \textsc{Neuron-Opsd} improves self-distillation while better preserving cross-domain performance and reducing calibration collapse.
\end{itemize}

\section{Related Work}

\paragraph{Annotation-free LLM Self-improvement.}

Recent work studies whether LLMs can improve without human labels by deriving supervision from their own outputs or intrinsic signals. Early self-improvement methods bootstrap rationales or self-generated feedback~\citep{zelikman_star_2022,yuan2024self, huang-etal-2023-large}, while recent annotation-free RL methods construct rewards from rollout agreement, entropy, or self-certainty~\citep{zuo_ttrl_2025,zhao_learning_2025}. Other lines explore self-play or zero-data training regimes~\citep{zhao2025absolute}. However, recent analysis of unsupervised RLVR shows that intrinsic rewards tend to sharpen the model's initial distribution and can collapse when confidence is misaligned with correctness~\citep{he_how_2026}. Our work follows this motivation but replaces hard output-level rewards with OPD, using neuron-derived signals to select data and construct teacher contexts without external annotations.

\textbf{LLM Probing through Internal Neuron Analysis.}
Internal-state analysis provides a way to inspect model behavior beyond surface outputs. Early probing methods such as the Logit Lens~\citep{nostalgebraist2020interpreting} and Tuned Lens~\citep{belrose2023eliciting} map intermediate representations into vocabulary space to study how predictions emerge across layers. More recent mechanistic work uses Sparse Autoencoders to disentangle superposed features and obtain more interpretable activation patterns~\citep{gao2024scaling}. Beyond interpretation, internal activations have also been linked to model reliability. Neuron agreement can signal when models are likely to be correct or hallucinate~\citep{chen_llms_2025, cao_model_2025}, and mechanism-interpretable metrics have been proposed to evaluate utility beyond surface accuracy~\citep{cao2025modelutilitylawevaluating}.  
Recent work has demonstrated a successful application of neuron-based signals to active few-shot learning, where neuron representations guide the selection of informative examples for annotation, improving few-shot performance with only a small set of high-quality annotated demonstrations~\cite{chen-etal-2026-neuron}.
Building on these findings, we extend the neuron-based data selection pipeline to reasoning models in the context of post-training, using neuron activation patterns as label-free signals to guide data selection and context construction for annotation-free self-improvement via on-policy distillation.

\textbf{On-Policy Distillation for LLMs.}
OPD trains a student on its own generated trajectories using token-level supervision from a teacher distribution~\citep{ye_-policy_2026,song2026surveyonpolicydistillationlarge}. This differs from standard off-policy distillation, where the student learns from fixed teacher-generated data, and from outcome-reward RL, where supervision is usually a scalar reward. Recent work has studied OPD for LLM post-training and self-distillation under settings with stronger teachers, privileged information, or context-conditioned teachers~\citep{hubotter_reinforcement_2026,zhao_self-distilled_2026,li2026rethinking}. Our work take OPD as the training objective, and focuses on the annotation-free setting where the teacher context and training data must be constructed without external labels.

\section{Preliminaries}

To establish the foundation for \textsc{N-OPSD}, we formalize the annotation-free self-evolution setting and the mechanics of OPD~\citep{ye_-policy_2026,hubotter_reinforcement_2026}. 

\subsection{Problem Formulation}
\label{sec:problem_formulation}

We consider an annotation-free self-evolution scenario. Given a base LLM with parameters $\theta$ and a domain-specific unlabeled dataset $\mathcal{D} = \{x_i\}_{i=1}^N$ of input queries, the objective is to improve the base model's performance on the target domain without access to ground-truth labels or any external model, i.e a different stronger teacher model.
We frame this as annotation-free post-training from a fixed unlabeled data pool: the only source of new signal during training is the model's own on-policy rollouts and intrinsic activation patterns. No environment interaction, human feedback, or external oracle is available. This is distinct from offline RL setups, which learn from logged reward-labeled trajectories, and from online RL post-training such as RLHF, which assumes access to live preference or reward queries.

\subsection{On-Policy Distillation}
On-Policy Distillation~\citep{song2026surveyonpolicydistillationlarge} trains the base or student model $\pi_\theta$ by minimizing a per-token reverse KL loss to a teacher model $\pi_{\text{teacher}}$ on trajectories sampled from the student model:
\begin{equation}
    \mathcal{L}_{\text{OPD}}(\theta) = \mathbb{E}_{\,x \sim \mathcal{D},\; y \sim \pi_\theta(\cdot \mid x)}\!\left[\sum_{t=1}^{|y|} \mathrm{KL}\!\left(\pi_\theta(\cdot \mid x, y_{<t}) \,\big\|\, \pi_{\text{teacher}}(\cdot \mid x, y_{<t})\right)\right],
    \label{eq:opd}
\end{equation}
The effectiveness of the student model largely depends on the quality of $\pi_{\text{teacher}}$, which is typically a larger model or the same model augmented with additional context, e.g., conditioning on external feedback like compiler traces~\citep{hubotter_reinforcement_2026}.
In our annotation-free setting, both teacher and student model are initialized from the same base LLM and no external feedback is available, so constructing a reliable and high-quality $\pi_{\text{teacher}}$ becomes the key challenge. To mitigate this, we condition the teacher model on auxiliary context $c$, which is constructed from unlabeled domain data alone, to improve its next-token distribution on the target task, i.e., $\pi_{\text{teacher}}(\cdot \mid x, y_{<t}) := \pi_\theta(\cdot \mid x, c, y_{<t})$. 


\subsection{Internal Signals for Data Selection}
\label{sec:pilot_study}

The additional context $c$ for the teacher model is constructed primarily through few-shot examples with the model's own generated solutions. Rather than using external supervision to verify the correctness of these pseudolabels, we probe the model's internal dynamics as a proxy for correctness, and the feature of the reasoning path. Now we introduce the calculation for the active neuron set.

\paragraph{Internal Neuron Signal Extraction.}
Building on the logit-lens view that intermediate representations can be read through the LM head~\citep{nostalgebraist2020interpreting}, and recent evidence that neuron-level activation features reflect response reliability~\citep{cao_model_2025,chen2025llms,chen-etal-2026-neuron}, we follow~\citet{chen-etal-2026-neuron} to extract the activated MLP neurons detailed as follows.
For layer $l$, let $\mathbf{h}^l$ be the MLP input, $\mathbf{W}_{\mathrm{in}}^l$ and $\mathbf{W}_{\mathrm{out}}^l$ be the up- and down-projection matrices. We compute the activation as:
\begin{equation}
    \mathbf{k}^l = \sigma(\mathbf{h}^l\mathbf{W}_{\mathrm{in}}^l),
\end{equation}
Next, given a generated token $\hat{y}$, we score each neuron at layer $l$ and index $i$ by projecting its downstream contribution to the unembedding space:
\begin{equation}
    S_{\hat{y}, i}^l = k_i^l \cdot (\mathbf{w}_{\mathrm{out}, i}^l \cdot \mathbf{e}_{\hat{y}}),
    \label{eq:contribution}
\end{equation}
where $\mathbf{w}_{\mathrm{out},i}^l$ is the $i$-th row of $\mathbf{W}_{\mathrm{out}}^l$ and $\mathbf{e}_{\hat{y}}$ is the unembedding vector for $\hat{y}$.
This early-unembedding score filters for neurons that actively promote the model's generated token.
For each generated token $y_i$, we define the activated neurons by keeping the top-K\footnote{We follow~\citet{chen-etal-2026-neuron} to set $K$ as 5000.} scored neurons across all layers. 
We then aggregate the retained neurons across the generated tokens in the response by taking their union set.
The resulting sparse set, denoted as $\mathcal{N}(x)$, serves as the internal signal used later for consensus estimation and context retrieval.

\section{Methodology}

We present a neuron-activation-guided framework for selecting training data for self-distillation and curate contexts for self-evolving LLMs.
Given an unlabeled data pool $\mathcal{D} = \{x_1, x_2, \ldots, x_N\}$, our approach selects a subset $\mathcal{S} \subset \mathcal{D}$ for self-distillation training, without requiring ground-truth labels.
The framework consists of four stages: neuron activation extraction, sample selection based on neuron-consensus, context curation via Jaccard-nearest few-shot retrieval, and self-evolving training.

\subsection{Neuron Activation Extraction}
\label{sec:neuron_extraction}
We instantiate the internal signal defined in \S\ref{sec:pilot_study} for every sample in the unlabeled pool.
Concretely, we collect rollouts with the model in zero-shot, compute early-unembedding contribution scores for MLP neurons, and store the set of activated neurons $\mathcal{N}(x)$ as the sample's sparse activation pattern.

\subsection{Sample Selection via Neuron Consensus}
\label{sec:selection}

Given the activation pattern $\mathcal{N}(x)$, we measure {neuron consensus} of $x$ by the size of its activation set:
\begin{equation}
    s(x) = |\mathcal{N}(x)|,
    \label{eq:consensus}
\end{equation}
where a smaller $s(x)$ indicates higher consensus, and a larger $s(x)$ indicates diffuse activation that could related to hallucinations~\cite{chen_llms_2025, chen-etal-2026-neuron}.
We interpret $s(x)$ as an annotation-free proxy for the model's sample-level reliability. Since samples with lower $s(x)$ tend to be answered more correctly, applying OPD-based self-improvement to these samples helps reinforce the model's already-correct reasoning and predictions, helping to avoid exacerbating hallucinations.

\subsection{Neuron Overlap for Teacher Context Construction}
\label{sec:jaccard}

To construct relevant few-shot contexts for the self-distillation teacher, we measure the overlap between neuron activation patterns using the Jaccard distance:
\begin{equation}
    J(x_i, x_j) = 1 - \frac{|\mathcal{N}(x_i) \cap \mathcal{N}(x_j)|}{|\mathcal{N}(x_i) \cup \mathcal{N}(x_j)|},
    \label{eq:jaccard}
\end{equation}
A small Jaccard distance indicates that the model processes two samples using similar neurons, suggesting they share similar knowledge circuits or go over similar reasoning path.
For each query sample $x_q$, we construct its $m$-shot demonstration set by selecting the $m$ nearest neighbors by:
\begin{equation}
    \mathcal{C}_K(x_q) = \underset{\substack{S \subset \mathcal{D} \setminus \{x_q\} \\ |S| = m}}{\arg\min} \sum_{x_j \in S} J(x_q, x_j),
    \label{eq:fewshot}
\end{equation}

This selection ensures that few-shot examples are processed through similar neural pathways as the query, providing the teacher with the most relevant contextual knowledge for distillation, helping sharpening its prediction distribution.

\subsection{Self-Improvement via On-Policy Distillation}
\label{sec:self_opd}

To enable annotation-free self-improvement, we integrate neuron-guided sample selection and context curation with OPD. 
The student is the zero-shot policy parameterized by the current model $\theta$, while the teacher is an EMA-updated copy $\bar{\theta}$ conditioned on the additional context $\mathcal{C}_K(x)$:
\begin{gather}
    \pi_{\theta}^{\mathrm{stu}}(\cdot \mid x, y_{<t}) 
    := \pi_\theta(\cdot \mid x, y_{<t}), \label{eq:student}\\
    q_{\bar{\theta}}^{\mathrm{tea}}(\cdot \mid \mathcal{C}_K(x), x, y_{<t}) 
    := \pi_{\bar{\theta}}(\cdot \mid \mathcal{C}_K(x), x, y_{<t}),
    \label{eq:teacher}\\
    \bar{\theta} \leftarrow (1-\tau)\, \bar{\theta} + \tau \theta,
\end{gather}
where $\mathcal{C}_K(x)$ denotes the additional context guided by neuron-overlap retrieved for sample $x$. 
During optimization, the teacher distribution is treated as a fixed target, and gradients are applied only to the student policy $\pi_{\theta}^{\mathrm{stu}}$.
The teacher is updated by exponential moving average (EMA) rate $\tau$.

The OPD training objective minimizes the reverse KL divergence on student-generated rollouts:
\begin{equation}
    \mathcal{L}_{\mathrm{OPD}}
    =
    \mathbb{E}_{x \sim \mathcal{S},\, y_{<t} \sim \pi_{\theta}^{\mathrm{stu}}}
    \left[
    \sum_{t=1}^{T}
    D_{\mathrm{KL}}\!\left(
        \pi_{\theta}^{\mathrm{stu}}(\cdot \mid x, y_{<t})
        \;\|\;
        q_{\bar{\theta}}^{\mathrm{tea}}(\cdot \mid c_x, x, y_{<t})
    \right)
    \right].
    \label{eq:opcd}
\end{equation}
where $\mathcal{S}$ denotes the selected subset, and $c_x=\mathcal{C}_K(x)$ is the neuron-overlap context retrieved for sample $x$. During optimization, the teacher distribution is treated as a fixed target, and gradients are applied only to the student policy. The teacher parameters $\bar{\theta}$ are updated by exponential moving average of the student parameters, yielding a temporally smoothed model updates.

This objective transfers the distributional benefits of neuron-aligned few-shot contexts into the zero-shot student policy, internalizing context-induced distributions through parameter updates and eliminating reliance on demonstrations at inference. Crucially, the entire pipeline is annotation-free: neuron activations guide both sample selection and context curation, while \textsc{N-OPSD} uses the model's own context-conditioned predictions as supervision. Algorithm~\ref{alg:nopsd} summarizes the full procedure.

\section{Experiments}
To establish the impact of each proposed component, we first analyze the effectiveness of each individually, namely, neuron consensus, neuron overlap, and the OPD training objective when combined with neuron consensus and neuron overlap. For the analyses in Section~\ref{analysis:neuron_con} and Section~\ref{analysis:neuron_overlap}, we use \textbf{SciKnowEval}~\citep{feng2025sciknowevalevaluatingmultilevelscientific}, a multiple-choice benchmark spanning four scientific domains, \textit{Biology}, \textit{Material}, \textit{Physics}, and \textit{Chemistry}, each split into 80\% train and 20\% test sets.
\subsection{Analysis: Neuron Consensus based Data Selection}
\label{analysis:neuron_con}

\begin{wrapfigure}{r}{5.5cm}
    \centering
    \includegraphics[width=5.5cm]{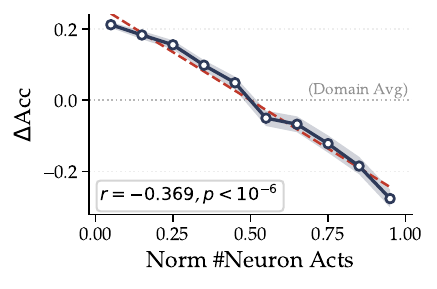}
    \caption{Neuron consensus correlates accuracy. $\Delta$Acc is the gap between average domain accuracy and bin accuracy, while \#Neuron Acts is normalized domain-wise. Separate domain-wise relational results shown in Figure~\ref{fig:domain-neu-hall-rel}.}
    \label{fig:acc_neu_rel}
\end{wrapfigure}

\textbf{Does Neuron Consensus Correlate with Performance?}
We validate the relationship between the activation density, i.e., the total count of activated neurons $|\mathcal{N}(x)|$, and the model's hallucination rates. 
As illustrated in Figure~\ref{fig:acc_neu_rel}, our analysis across all four SciKnowEval domains reveals a strong, robust monotonic correlation: queries that activate a larger number of neurons are significantly more likely to result in incorrect rollouts.

This phenomenon is attributed to {Neuron Consensus}. Sparser activations with lower $|\mathcal{N}(x)|$ indicate a higher degree of consensus within the network. The model effortlessly retrieves a clear, unconflicted knowledge trace. Conversely, a denser, widespread activation pattern suggests internal conflict and uncertainty, where the model struggles to reconcile disparate neural pathways. This finding fundamentally addresses our data selection bottleneck: by isolating queries with the lowest activation counts, we can filter out hallucination-prone samples and secure pristine targets for self-supervision, even without external labels.

\textbf{Are Easy Samples Better Training Data for OPD Training?}
We rank the source pool by neuron count $\mathrm{s(x)}$ and train on the subset that ranked top and bottom 20\% following the OPD framework. For evaluation, we sample 8 rollouts for each test query and report two metrics: \textbf{Avg@8}, the mean per-sample accuracy; and \textbf{ECE}, the expected calibration error over the per-query majority-vote confidence with 15 equal-width bins, lower the better.
Table.~\ref{tab:ablation_selection} reports the results on SciKnowEval. 
Intuitively, subsets with higher activation numbers, i.e. Top-20\%, will generally noisier than the subsets with lower activation numbers, i.e., Bottom-20\%. As lower activation numbers indicates a lower hallucination rates. 
However, in training ablations, with OPD, Top-20\% can still have reasonable improvements on some domains such as Bio. and Mat., also with a competitive calibration.

We hypothesize that this is due to the trade-off between learning utility and noise levels. Specifically, samples prone to hallucination could offer higher learning value since they represent areas where the LLM consistently fails. However, in self-improvement scenarios, the model struggles to generate reliable supervision signals for these difficult cases. Conversely, while the LLM can easily produce correct self-generated signals for simpler samples, their contribution to overall improvement is marginal. Given this persistent trade-off, training on a refined top-20\% subset can still yield substantial performance gains, even in the presence of noise. 

\begin{table*}[h]
\centering
\setlength{\tabcolsep}{3pt}
\renewcommand{\arraystretch}{1}

\scalebox{0.75}{
\begin{tabular}{lcccccccc}
\toprule
\multirow{2}{*}{\textbf{Selection}} & \multicolumn{4}{c}{\textbf{Avg@8} ($\uparrow$)} & \multicolumn{4}{c}{\textbf{ECE} ($\downarrow$)} \\
\cmidrule(lr){2-5} \cmidrule(lr){6-9}
 & \textbf{\textsc{Chem.}} & \textbf{\textsc{Bio.}} & \textbf{\textsc{Mat.}} & \textbf{\textsc{Phys.}} & \textbf{\textsc{Chem.}} & \textbf{\textsc{Bio.}} & \textbf{\textsc{Mat.}} & \textbf{\textsc{Phys.}} \\
\midrule
{Qwen3-4B} & 72.11 & 73.98 & 71.12 & 80.20 & 0.173 & 0.195 & 0.213 & 0.095 \\
\midrule
+Top-20\%    & 68.46$_{\textcolor{negred}{-3.65}}$ & \textbf{74.59}$_{\textcolor{posgreen}{+0.61}}$ & \textbf{73.82}$_{\textcolor{posgreen}{+2.70}}$ & 82.00$_{\textcolor{posgreen}{+1.80}}$ & \textbf{0.133}$_{\textcolor{posgreen}{-0.040}}$ & 0.198$_{\textcolor{negred}{+0.003}}$ & \textbf{0.194}$_{\textcolor{posgreen}{-0.019}}$ & 0.116$_{\textcolor{negred}{+0.021}}$ \\
+\textbf{Bottom-20\%} & \textbf{72.03}$_{\textcolor{negred}{-0.08}}$ & 74.02$_{\textcolor{posgreen}{+0.04}}$ & 73.39$_{\textcolor{posgreen}{+2.27}}$ & \textbf{83.13}$_{\textcolor{posgreen}{+2.93}}$ & 0.171$_{\textcolor{posgreen}{-0.002}}$ & \textbf{0.193}$_{\textcolor{posgreen}{-0.002}}$ & 0.207$_{\textcolor{posgreen}{-0.006}}$ & \textbf{0.114}$_{\textcolor{negred}{+0.019}}$ \\
\bottomrule
\end{tabular}
}
\caption{
Analysis results of consensus-based selection, reported in-domain Avg@8 and ECE. 
}
\label{tab:ablation_selection}
\end{table*}


\subsection{Analysis: Neuron Overlap Guided Context Curation}
\label{analysis:neuron_overlap}

In this experiment, we fix the Bottom-20\% as OPD training data, we compare neuron-overlap retrieval with uniform in-domain random retrieval. As shown in Table.~\ref{tab:ablation_retrieval}, \textsc{Neuron-Jaccard} improves over the base model on three of the four SciKnowEval sources and outperforms random retrieval on \textsc{Chem.} and \textsc{Phys.}. Random in-domain retrieval is also competitive on \textsc{Bio.} and \textsc{Mat.}, indicating that in-domain demonstrations already provide a strong context prior.
\begin{table*}[h]
\centering

\renewcommand{\arraystretch}{.8}

\scalebox{0.75}{
\begin{tabular}{lcccc}
\toprule
\textbf{Retriever} & \textbf{\textsc{Chem.}} & \textbf{\textsc{Bio.}} & \textbf{\textsc{Mat.}} & \textbf{\textsc{Phys.}} \\
\midrule
{Qwen3-4B-Thinking} & 72.11 & 73.98 & 71.12 & 80.20 \\
\midrule
+Random & 68.48$_{\textcolor{negred}{-3.63}}$ & \textbf{74.14}$_{\textcolor{posgreen}{+0.16}}$ & \textbf{73.79}$_{\textcolor{posgreen}{+2.67}}$ & 81.97$_{\textcolor{posgreen}{+1.77}}$ \\
+\textbf{Neuron-Jaccard (ours)} & \textbf{72.03}$_{\textcolor{negred}{-0.08}}$ & 74.02$_{\textcolor{posgreen}{+0.04}}$ & 73.39$_{\textcolor{posgreen}{+2.27}}$ & \textbf{83.13}$_{\textcolor{posgreen}{+2.93}}$ \\
\bottomrule
\end{tabular}
}
\caption{Analysis results of the few-shot retriever, reported in-domain Avg@8 on SciKnowEval.}
\label{tab:ablation_retrieval}
\end{table*}

We further analyze why the marginal gain over random varies across domains. Because both random and neuron-based retrieval are restricted to the same source domain, the difference depends on how discriminative the within-domain contexts are. We measure reasoning diversity by the mean pairwise neuron-Jaccard distance between training samples: \textsc{Chem.} $0.870$, \textsc{Bio.} $0.634$, \textsc{Mat.}\ $0.628$, and \textsc{Phys.}\ $0.640$. In more diverse domains, such as \textsc{Chem.}, neuron-overlap retrieval can select demonstrations that match the query-specific reasoning mode, yielding a clear advantage over random selection. In more homogeneous domains, many examples activate similar neuron sets, so nearest-neighbor and random in-domain contexts become partially redundant, reducing the marginal benefit of Jaccard retrieval. This explains why random retrieval remains competitive on \textsc{Bio.} and \textsc{Mat.}, while also showing that \textsc{Neuron Overlap} is most useful when neuron similarity can distinguish meaningful reasoning modes within the source pool.

\subsection{Experimental Setup}
We then perform the evaluations on various self-improvement training algorithms. As we didn't observe a significant performance and calibration difference of using the subsets with the bottom 20\% vs the top 20\% \#neuron activations, therefore we demonstrated the bottom 20\% results. 

\textbf{Datasets.}
We evaluate on three different datasets. \textbf{SciKnowEval}~\citep{feng2025sciknowevalevaluatingmultilevelscientific} contains four scientific multiple-choice domains: \textit{Biology}, \textit{Material}, \textit{Physics}, and \textit{Chemistry}, each split 80/20 for training and testing.
\textbf{Edu-Feedback}~\citep{wu2023passive} is a binary feedback-quality classification dataset with 1,799 training and 1,000 testing samples.
\textbf{MMLU-Pro}~\citep{wang2024mmlu} is a challenging multi-domain multiple-choice benchmark. We split 80\% data for self-improvement.

\textbf{Baselines and Models.}
We compare with both the off-policy and on-policy self-improvement baselines. Including SFT-based \textbf{LMSI}~\citep{huang-etal-2023-large}, on-policy methods \textbf{TTRL}~\citep{zuo_ttrl_2025}, and \textbf{Intuitor}~\citep{zhao_learning_2025}, an annotation-free RL method using self-certainty as reward. 
We evaluate on {Qwen3-4B-Thinking-2507}~\citep{yang2025qwen3}, abbreviated as Qwen-4B in our paper.

\textbf{Metrics.}
We sample 8 rollouts for each test query and report three metrics: \textbf{Avg@8}, the mean per-sample accuracy; \textbf{Maj@8}, the majority-vote accuracy; and \textbf{ECE}, the expected calibration error over the per-query majority-vote confidence with 15 equal-width bins, lower the better.

\subsection{Performance Results}
\label{sec:OPSD_vs_ttrl}

To more comprehensively evaluate the efficacy of the self-improving baselines and the proposed method, we focus on both in-domain and cross-domain performance gain and calibration variances.
Specifically, we train each method on a single source 
domain and evaluate on all target test sets.
Table.~\ref{tab:cross_domain_compressed} and Table.~\ref{tab:cross_domain_ece_compressed} report Accuracy and ECE on Qwen3-4B-Thinking-2507. 


\definecolor{posgreen}{rgb}{0.0, 0.55, 0.0}
\definecolor{negred}{rgb}{0.75, 0.0, 0.0}

\begin{table*}[h!]
\centering
\small
\setlength{\tabcolsep}{4.5pt}
\renewcommand{\arraystretch}{.8}
\newcolumntype{C}{>{\centering\arraybackslash}p{13mm}}

\scalebox{0.86}{
\begin{tabular}{llCCCCCCCC}
\toprule
\multirow{2}{*}{\textbf{Method}} 
& \multirow{2}{*}{\textbf{Eval.}} 
& \multicolumn{5}{c}{\textbf{SciKnowEval}} 
& \multirow{2}{*}{\textbf{Edu.}} 
& \multirow{2}{*}{\makecell{\textbf{MMLU-}\\\textbf{Pro}}} 
& \multirow{2}{*}{\textbf{Avg.}} \\
& & \textsc{Bio} & \textsc{Mat.} & \textsc{Phys.} & \textsc{Chem} & \textit{Avg.} & & & \\
\midrule
\multicolumn{2}{l}{Qwen3-4B base}
& 73.98 & 71.12 & 80.20 & 72.11 & 74.35 & 64.95 & 72.16 & 72.42 \\
\midrule

\multirow{2}{*}{LMSI}
& in
& 74.59$_{\textcolor{posgreen}{+0.61}}$
& 69.08$_{\textcolor{negred}{-2.04}}$
& 78.54$_{\textcolor{negred}{-1.66}}$
& 73.72$_{\textcolor{posgreen}{+1.61}}$
& 73.98$_{\textcolor{negred}{-0.37}}$
& 66.95$_{\textcolor{posgreen}{+2.00}}$
& 68.60$_{\textcolor{negred}{-3.56}}$
& 71.91$_{\textcolor{negred}{-0.51}}$ \\
& cross
& 68.57$_{\textcolor{negred}{-3.54}}$
& 65.48$_{\textcolor{negred}{-7.20}}$
& 67.02$_{\textcolor{negred}{-3.84}}$
& 71.19$_{\textcolor{negred}{-1.29}}$
& 68.06$_{\textcolor{negred}{-3.97}}$
& 58.14$_{\textcolor{negred}{-15.77}}$
& 73.46$_{\textcolor{posgreen}{+0.99}}$
& 67.31$_{\textcolor{negred}{-5.11}}$ \\
\midrule

\multirow{2}{*}{TTRL}
& in
& 75.00$_{\textcolor{posgreen}{+1.02}}$
& 73.26$_{\textcolor{posgreen}{+2.14}}$
& 82.98$_{\textcolor{posgreen}{+2.78}}$
& 72.43$_{\textcolor{posgreen}{+0.32}}$
& 75.92$_{\textcolor{posgreen}{+1.56}}$
& 64.95$_{\textcolor{posgreen}{+0.00}}$
& 72.04$_{\textcolor{negred}{-0.12}}$
& 73.44$_{\textcolor{posgreen}{+1.02}}$ \\
& cross
& \textbf{73.31}$_{\textcolor{posgreen}{+1.20}}$
& \textbf{73.32}$_{\textcolor{posgreen}{+0.64}}$
& 71.28$_{\textcolor{posgreen}{+0.41}}$
& 73.46$_{\textcolor{posgreen}{+0.97}}$
& \textbf{72.84}$_{\textcolor{posgreen}{+0.81}}$
& 74.47$_{\textcolor{posgreen}{+0.55}}$
& \textbf{73.76}$_{\textcolor{posgreen}{+1.29}}$
& 73.27$_{\textcolor{posgreen}{+0.84}}$ \\
\midrule

\multirow{2}{*}{Intuitor\footnotemark}
& in
& \textbf{76.21}$_{\textcolor{posgreen}{+2.23}}$
& \textbf{73.41}$_{\textcolor{posgreen}{+2.29}}$
& \textbf{83.43}$_{\textcolor{posgreen}{+3.23}}$
& \textbf{72.63}$_{\textcolor{posgreen}{+0.52}}$
& \textbf{76.42}$_{\textcolor{posgreen}{+2.07}}$
& ---
& 71.91$_{\textcolor{negred}{-0.25}}$
& 75.52$_{\textcolor{posgreen}{+1.60}}$ \\
& cross
& 69.50$_{\textcolor{negred}{-2.61}}$
& 72.79$_{\textcolor{posgreen}{+0.11}}$
& 68.33$_{\textcolor{negred}{-2.53}}$
& 70.76$_{\textcolor{negred}{-1.72}}$
& 70.35$_{\textcolor{negred}{-1.69}}$
& ---
& 73.30$_{\textcolor{posgreen}{+0.83}}$
& 70.94$_{\textcolor{negred}{-1.18}}$ \\
\midrule

\multirow{2}{*}{N-OPSD}
& in
& 74.02$_{\textcolor{posgreen}{+0.04}}$
& 73.39$_{\textcolor{posgreen}{+2.27}}$
& 83.13$_{\textcolor{posgreen}{+2.93}}$
& 72.03$_{\textcolor{negred}{-0.08}}$
& 75.64$_{\textcolor{posgreen}{+1.29}}$
& \textbf{72.19}$_{\textcolor{posgreen}{+7.24}}$
& 72.04$_{\textcolor{negred}{-0.12}}$
& \textbf{74.47}$_{\textcolor{posgreen}{+2.05}}$ \\
& cross
& 72.71$_{\textcolor{posgreen}{+0.61}}$
& 73.27$_{\textcolor{posgreen}{+0.59}}$
& \textbf{71.37}$_{\textcolor{posgreen}{+0.51}}$
& \textbf{73.59}$_{\textcolor{posgreen}{+1.11}}$
& 72.74$_{\textcolor{posgreen}{+0.70}}$
& \textbf{75.22}$_{\textcolor{posgreen}{+1.31}}$
& 73.74$_{\textcolor{posgreen}{+1.27}}$
& \textbf{73.32}$_{\textcolor{posgreen}{+0.90}}$ \\
\bottomrule
\end{tabular}
}

\caption{
Cross-domain evaluation on Qwen3-4B, Avg@8.
For each method, \textit{in} reports the score of the model trained and evaluated on the same source domain, while \textit{cross} reports the average score of the same source-trained model on the other target domains.
Subscripts report $\Delta$ against the untrained Qwen3-4B baseline computed on the same target-domain support.
The full results are in Table~\ref{tab:cross_domain_main_avg8}.
}
\label{tab:cross_domain_compressed}
\end{table*}
\footnotetext{The Intuitor training on the Edu. dataset yields no improvement across any validation steps, thus we drop the results.}

\begin{table*}[h!]
\centering
\small
\setlength{\tabcolsep}{6pt}
\renewcommand{\arraystretch}{.8}
\newcolumntype{C}{>{\centering\arraybackslash}p{13mm}}

\scalebox{0.82}{
\begin{tabular}{llCCCCCCCC}
\toprule
\multirow{2}{*}{\textbf{Method}} 
& \multirow{2}{*}{\textbf{Eval.}} 
& \multicolumn{5}{c}{\textbf{SciKnowEval}} 
& \multirow{2}{*}{\textbf{Edu.}} 
& \multirow{2}{*}{\makecell{\textbf{MMLU-}\\\textbf{Pro}}} 
& \multirow{2}{*}{\textbf{Avg.}} \\
& & \textsc{Bio} & \textsc{Mat.} & \textsc{Phys.} & \textsc{Chem} & \textit{Avg.} & & & \\
\midrule
\multicolumn{2}{l}{Qwen3-4B base}
& 0.195 & 0.213 & 0.095 & 0.173 & 0.169 & 0.246 & 0.184 & 0.184 \\
\midrule

\multirow{2}{*}{LMSI}
& in
& 0.202$_{\textcolor{negred}{+0.007}}$
& 0.225$_{\textcolor{negred}{+0.012}}$
& \textbf{0.111}$_{\textcolor{negred}{+0.016}}$
& 0.179$_{\textcolor{negred}{+0.006}}$
& 0.179$_{\textcolor{negred}{+0.010}}$
& 0.255$_{\textcolor{negred}{+0.009}}$
& \textbf{0.171}$_{\textcolor{posgreen}{-0.013}}$
& 0.191$_{\textcolor{negred}{+0.007}}$ \\
& cross
& \textbf{0.171}$_{\textcolor{posgreen}{-0.011}}$
& 0.179$_{\textcolor{negred}{+0.001}}$
& 0.202$_{\textcolor{negred}{+0.000}}$
& \textbf{0.182}$_{\textcolor{posgreen}{-0.004}}$
& \textbf{0.184}$_{\textcolor{posgreen}{-0.004}}$
& \textbf{0.095}$_{\textcolor{posgreen}{-0.077}}$
& \textbf{0.162}$_{\textcolor{posgreen}{-0.022}}$
& \textbf{0.165}$_{\textcolor{posgreen}{-0.019}}$ \\
\midrule

\multirow{2}{*}{TTRL}
& in
& 0.204$_{\textcolor{negred}{+0.009}}$
& 0.208$_{\textcolor{posgreen}{-0.005}}$
& 0.123$_{\textcolor{negred}{+0.028}}$
& 0.192$_{\textcolor{negred}{+0.019}}$
& 0.182$_{\textcolor{negred}{+0.013}}$
& 0.248$_{\textcolor{negred}{+0.002}}$
& 0.198$_{\textcolor{negred}{+0.014}}$
& 0.196$_{\textcolor{negred}{+0.012}}$ \\
& cross
& 0.186$_{\textcolor{negred}{+0.004}}$
& 0.187$_{\textcolor{negred}{+0.009}}$
& 0.204$_{\textcolor{negred}{+0.002}}$
& 0.199$_{\textcolor{negred}{+0.012}}$
& 0.194$_{\textcolor{negred}{+0.007}}$
& 0.178$_{\textcolor{negred}{+0.006}}$
& 0.190$_{\textcolor{negred}{+0.006}}$
& 0.191$_{\textcolor{negred}{+0.007}}$ \\
\midrule

\multirow{2}{*}{Intuitor}
& in
& 0.208$_{\textcolor{negred}{+0.013}}$
& 0.212$_{\textcolor{posgreen}{-0.001}}$
& 0.117$_{\textcolor{negred}{+0.022}}$
& 0.198$_{\textcolor{negred}{+0.026}}$
& 0.184$_{\textcolor{negred}{+0.015}}$
& ---
& 0.174$_{\textcolor{posgreen}{-0.010}}$
& 0.182$_{\textcolor{negred}{+0.010}}$ \\
& cross
& 0.226$_{\textcolor{negred}{+0.044}}$
& 0.187$_{\textcolor{negred}{+0.008}}$
& 0.234$_{\textcolor{negred}{+0.032}}$
& 0.224$_{\textcolor{negred}{+0.038}}$
& 0.218$_{\textcolor{negred}{+0.030}}$
& ---
& 0.185$_{\textcolor{negred}{+0.000}}$
& 0.211$_{\textcolor{negred}{+0.024}}$ \\
\midrule

\multirow{2}{*}{N-OPSD}
& in
& \textbf{0.193}$_{\textcolor{posgreen}{-0.003}}$
& \textbf{0.207}$_{\textcolor{posgreen}{-0.006}}$
& 0.114$_{\textcolor{negred}{+0.019}}$
& \textbf{0.171}$_{\textcolor{posgreen}{-0.002}}$
& \textbf{0.171}$_{\textcolor{negred}{+0.002}}$
& \textbf{0.191}$_{\textcolor{posgreen}{-0.056}}$
& 0.180$_{\textcolor{posgreen}{-0.004}}$
& \textbf{0.176}$_{\textcolor{posgreen}{-0.008}}$ \\
& cross
& 0.177$_{\textcolor{posgreen}{-0.005}}$
& \textbf{0.177}$_{\textcolor{posgreen}{-0.001}}$
& \textbf{0.194}$_{\textcolor{posgreen}{-0.008}}$
& 0.187$_{\textcolor{negred}{+0.000}}$
& 0.184$_{\textcolor{posgreen}{-0.004}}$
& 0.177$_{\textcolor{negred}{+0.005}}$
& 0.186$_{\textcolor{negred}{+0.002}}$
& 0.183$_{\textcolor{posgreen}{-0.001}}$ \\
\bottomrule
\end{tabular}
}

\caption{
Cross-domain calibration on Qwen3-4B, reported as ECE, computed from majority-vote confidence.
For each method, \textit{in} denotes source-domain ECE, while \textit{cross} denotes the average ECE of the same source-trained model on other domains.
The full results are in Table~\ref{tab:cross_domain_ece_avg8}.
}
\label{tab:cross_domain_ece_compressed}
\end{table*}

\textbf{Performance.}
N-OPSD improves source-domain performance on four of the six training sources in Table.~\ref{tab:cross_domain_compressed}, with especially clear gains on \textsc{Mat.}, \textsc{Phys.}, and \textsc{Edu.}
On \textsc{Mat.}, \textsc{Phys.}, and \textsc{Edu.}, N-OPSD matches or exceeds TTRL's source-domain gain, and its averaged row attains the highest overall Avg@8 among the compared methods.
The main advantage is not uniform source-domain dominance, but a stronger accuracy-preservation trade-off: every SciKnow-trained N-OPSD model improves \textsc{Edu.} over the base model, whereas TTRL is mixed. LMSI and Intuitor often suffer large drops.
Thus, N-OPSD delivers source-domain improvement where the teacher provides useful sharpening while better preserving non-source capability.


\textbf{Calibration.}
Table.~\ref{tab:cross_domain_ece_compressed} shows that TTRL and Intuitor generally increase average ECE, while N-OPSD keeps the calibration cost smaller and slightly reduces overall ECE in the averaged row.
The strongest calibration effect appears on \textsc{Edu.}: every N-OPSD row lowers \textsc{Edu.}\ ECE relative to the base model, whereas TTRL is mixed and Intuitor often worsens it substantially.
Several N-OPSD rows simultaneously improve overall Avg@8 and reduce overall ECE, and N-OPSD$_{\textsc{Edu.}}$ gives the largest overall accuracy gain together with the lowest overall ECE among N-OPSD rows.
These results show where the method works best, it improves accuracy without the broad calibration inflation seen in reward-based annotation-free RL, although it can still increase ECE on some SciKnowEval targets.

\subsection{What drives the self-improvement, and when does it benefit most?}
\label{sec:iterative}





In terms of self-training, \citet{he_how_2026} characterise intrinsic self-supervision methods as sharpening the model's prior distribution rather than adding new information, which succeeds only when that prior is already aligned with correctness. Our consensus stratification makes the trade concrete at the sample level, with high-consensus samples supplying a reliable prior to sharpen against and low-consensus samples supplying an unreliable one. 
Meanwhile, OPD derives its training signal from a soft teacher-student distributional gap rather than from hard scalar rewards. This allows part of the teacher's uncertainty to be preserved during distillation, yielding a more favorable trade-off between calibration and correctness.
Here, we conduct a data-centric analysis of the selected training pool to examine how data dynamics drive the gains of OPD-based LLM self-training.

\begin{table*}[h]
\centering

\scalebox{.75}{
\begin{tabular}{lccccc}
\toprule
\textbf{Domain} & \textbf{Base Test Avg@8} & \textbf{Train Maj@8} & \textbf{Train Avg@8} & \textbf{Maj$-$Avg} & \textbf{in-domain gain} \\
\midrule
\textsc{Chem.}    & 72.11 & 98.23 & 98.18 & 0.04 & $-0.08$ \\
\textsc{Bio.}     & 73.98 & 98.73 & 98.64 & 0.09 & $+0.04$ \\
\textsc{Mat.}     & 71.12 & 91.93 & 91.59 & 0.34 & $+2.27$ \\
\textsc{Phys.}    & 80.20 & 96.31 & 96.27 & 0.04 & $+2.93$ \\
\bottomrule
\end{tabular}
}
\caption{Base rollout statistics on the training pool versus N-OPSD's in-domain Avg@8 gain. All statistics are computed with Qwen3-4B-Thinking-2507: Test Avg@8 is measured on the held-out test set, while training-pool statistics are measured on the bottom-20\% subset by neuron count.}
\label{tab:sharpening_room}
\end{table*}

\begin{table*}[h]
\centering
\scalebox{0.75}{
\begin{tabular}{lccccc}
\toprule
\textbf{Domain} & \textbf{$H_{\text{zs}}$ Student }& \textbf{$H_{\text{neur}}$ Teacher} & \textbf{$\Delta H$ T$-$S} &  \textbf{{Avg@8 Gain}} \\
\midrule
\textsc{Chem.}  & 0.240 & 0.237 & $-0.002$                   & $-0.08$ \\
\textsc{Bio.}   & 0.238 & 0.240 & $+0.001$                    & $+0.04$ \\
\textsc{Mat.}   & 0.268 & 0.251 & ${-0.017}$    & $+2.27$ \\
\textsc{Phys.}  & 0.257 & 0.247 & ${-0.010}$     & $+2.93$ \\
\bottomrule
\end{tabular}
}
\caption{Teacher-student per-token logprob entropy on the selected training pool. The teacher conditions on $K{=}10$ neuron-Jaccard nearest demonstrations, and the student is zero-shot. $\Delta H<0$ means the teacher is sharper at the token level. The two domains with significant teacher sharpening, \textsc{Mat.}\ and \textsc{Phys.}, are also the two with the largest in-domain Avg@8 gain.}
\label{tab:teacher_student_bot20}
\end{table*}

\textbf{Sharpening Room on High-Consensus Samples.}
A necessary precondition for N-OPSD to deliver a meaningful in-domain gain is that the high neuron-consensus pool used for distillation still contains enough gap for sharpening, i.e., carries disagreement across the rollouts.
Table.~\ref{tab:sharpening_room} reports the $n{=}8$ rollout stats of the untrained model on this pool. 
Across \textsc{Chem.}, \textsc{Bio.}, and \textsc{Mat.}, the Maj$-$Avg gap on the high-consensus training pool tracks N-OPSD's in-domain gain: larger residual rollout disagreement gives OPD distillation more uncertainty to resolve. However, \textsc{Phys.}\ is already near-unanimous and still obtains a large gain, showing that vote-level sharpening room alone is not sufficient to explain all improvements.


\textbf{Teacher-Student Gap Drives Gain.}
The sharpening-room analysis captures part of when N-OPSD has signal to exploit, but not whether the few-shot teacher provides a useful target. On the selected pool, vote-level metrics saturate, as both the zero-shot student and the neuron-Jaccard teacher reach ${\sim}98\%$ Avg@8 in Table.~\ref{tab:sharpening_room}, so the teacher-student gap must be measured at the token level. Table.~\ref{tab:teacher_student_bot20} reports per-token logprob entropy, averaged over the response trajectory and paired across queries. On \textsc{Mat.}\ and \textsc{Phys.}\ the teacher meaningfully sharpens the token distribution, with $\Delta H = -0.017$ and $-0.010$, and these are exactly the two domains where N-OPSD delivers the strongest in-domain gains of $+2.27$ and $+2.93$. 
On \textsc{Bio.}\ and \textsc{Chem.}\ the teacher provides no token-level sharpening, with $|\Delta H|<0.003$ and no significant difference, and the in-domain gain collapses correspondingly. 


\section{Conclusion}

In this work, we propose \textsc{Neuron-Opsd} for annotation-free post-training from a fixed unlabeled data pool, where no ground-truth annotations are available.
The method uses Neuron Consensus to select more reliable self-improvement data and Neuron Overlap to construct neuron-aligned teacher contexts for on-policy distillation.
These two components are proposed for the central challenge of unreliable self-supervision, which can otherwise amplify hallucinations, miscalibration, or negative transfer.
Across our evaluations, \textsc{Neuron-Opsd} achieves competitive in-domain gains while better preserving cross-domain capability and calibration than prior annotation-free methods.

\section*{Limitations}

There are several limitations remaining in this work.
First, in-domain gains are not uniform across domains: \textsc{Bio.} and \textsc{Chem.} show negligible improvement, consistent with Tab.~\ref{tab:teacher_student_bot20} showing no teacher token-distribution sharpening on those domains.
Second, although \textsc{Neuron Consensus} correlates with hallucinated behavior, our top- and bottom-activation ablations show that activation count alone is insufficient for data selection. This poses a fundamental question: how do we balance self-improvement against signal reliability? Identifying the precise conditions under which LLMs can be trusted to self-supervise is left for future work. Meanwhile, a deeper theoretical analysis of the neuron consensus and its correlation with other hallucination signals is needed for a proper usage of this signal, such as entropy, reasoning length, question complexity, etc.
Third, \textsc{Neuron Overlap}'s retrieval coherence depends on domain reasoning diversity; as the case study in App.~\ref{app:jaccard_case_study} shows, domains with homogeneous reasoning patterns can make neuron-Jaccard retrieval less discriminative and push contexts toward generic examples.
Fourth, although we compare against uniform in-domain random few-shot retrieval, we do not compare against other alternative teacher-context construction strategies, such as retrieval-augmented or prompt-engineered contexts, that might produce similar teacher sharpening without neuron overlap.
Also, the evaluation covers three datasets, SciKnowEval, Edu-Feedback, and MMLU-Pro, so the observed side-effect tradeoffs may change under broader domain coverage and different context-construction mechanisms. 
Meanwhile, the proposed method has only been evaluated on models of 4B parameters. Whether the observed benefits scale to larger LLMs remains unclear, and experiments on stronger model families are needed.

\bibliographystyle{plainnat}
\bibliography{custom,LLM-Self_improvement}

\appendix
\appendix

\section{Implementation Details}
\label{app:implementation}

\paragraph{Neuron contribution computation.}
We register forward hooks on the activation function of each transformer MLP layer, \texttt{model.layers[l].mlp.act\_fn}.
For each response token position $t$, we capture the post-activation hidden state $\mathbf{a}_{l,t} \in \mathbb{R}^{d_{\text{inter}}}$ and compute contribution scores via Eq.~\ref{eq:contribution}.
We retain the top 2,000 neurons per layer per chunk, then apply global Top-$K$ deduplication with $K = 5{,}000$ across all layers.
When the same layer-neuron pair appears across multiple chunks, namely groups of response positions, we keep the maximum contribution score.


\paragraph{N-OPSD training configuration.}
All N-OPSD models are trained using the veRL framework with Ray-based FSDP distributed training across 4 GPUs, batch size 16, micro-batch size 4, and 150 training steps. We use reverse KL divergence and EMA teacher updates with rate $0.01$. 


\paragraph{TTRL training configuration.}
TTRL models follow the standard open-source setup of \citet{zuo_ttrl_2025}, using GRPO with majority-vote pseudo-labels. We train with 4 GPUs, batch size 16, actor learning rate $5 \times 10^{-7}$ with cosine warmup and 3\% warmup ratio, critic learning rate $9 \times 10^{-6}$, KL coefficient $0.0$, rollout temperature $1.0$, and 8 votes per prompt. Validation uses temperature $0.6$ with 8 samples. Models are trained for 1 epoch with checkpoint selection based on the best validation accuracy, maj@8.

\paragraph{Data selection configuration.}
For the neuron-guided data selection used by N-OPSD, we apply the same rule across all domains: \textit{bot20}, which retains the bottom-20\% of the source pool by zero-shot sensitivity, where $s(x) = |\mathcal{N}_{0\text{-shot}}(x)|$ is the number of unique neurons activated without context. The selected pool is then paired with $K=10$ Jaccard-nearest few-shot demonstrations during training; see \S\ref{sec:jaccard}. No ground-truth labels are used at any point during selection.

\section{Methodology}

\subsection{Algorithm}
\begin{algorithm}[h!]
\DontPrintSemicolon
\KwIn{Unlabeled pool $\mathcal{D}$; LLM $\pi_\theta$; percentile $b$; shots $K$; training steps $T$; EMA rate $\rho$; learning rate $\eta$.}
\KwOut{Updated model $\pi_\theta$.}
\tcp{Stage 1: neuron activation extraction (\S\ref{sec:neuron_extraction})}
\ForEach{$x \in \mathcal{D}$}{
    Generate $y \sim \pi_\theta(\cdot \mid x)$ and score neurons via Eq.~\ref{eq:contribution}\;
    Form activation set $\mathcal{N}(x)$ and neuron count $s(x) = |\mathcal{N}(x)|$\;
}
\tcp{Stage 2: high-consensus selection (\S\ref{sec:selection})}
$\tau_b \leftarrow$ $b$-th percentile of $\{s(x)\}_{x \in \mathcal{D}}$\;
$\mathcal{S} \leftarrow \{x \in \mathcal{D} \mid s(x) \leq \tau_b\}$\;
\tcp{Stage 3: context construction (\S\ref{sec:jaccard})}
Precompute $J(x_i, x_j)$ for all pairs in $\mathcal{D}$ via Eq.~\ref{eq:jaccard}\;
\ForEach{$x \in \mathcal{S}$}{
    $\mathcal{C}_K(x) \leftarrow$ $K$-nearest neighbors of $x$ in $\mathcal{D}\setminus\{x\}$ under $J$\;
}
\tcp{Stage 4: self-training (\S\ref{sec:self_opd})}
Initialise EMA teacher $\theta' \leftarrow \theta$\;
\For{$t = 1,\ldots,T$}{
    Sample batch $B \subset \mathcal{S}$\;
    Generate student rollouts $y \sim \pi_\theta(\cdot \mid x)$ for $x \in B$\;
    Compute $\mathcal{L}_{\text{OPD}}$ between $\pi_\theta(\cdot \mid x)$ and $\pi_{\theta'}(\cdot \mid x, \mathcal{C}_K(x))$ via Eq.~\ref{eq:opcd}\;
    $\theta \leftarrow \theta - \eta \nabla_\theta \mathcal{L}_{\text{OPD}}$\;
    $\theta' \leftarrow (1-\rho)\,\theta' + \rho\,\theta$ \tcp*{EMA update}
}
\Return $\pi_\theta$\;
\caption{Neuron-OPSD}
\label{alg:nopsd}
\end{algorithm}

\section{Experiments}

\subsection{Evaluation}
All evaluations sample 8 responses per question with temperature $0.6$ using vLLM. We report majority-vote accuracy, maj@8, and Expected Calibration Error (ECE), where the confidence for each question is $c = \mathrm{maxcount}/8$.







\subsection{Computational Cost.}
Besides the distillation training cost, the additional computational overhead in the proposed framework stems from the neuron-based few-shot and sample selection process. Specifically, this process requires forward passes through the initial LLM to obtain neuron features, followed by computing the distance matrix and the corresponding dataset statistics over these features. However, the entire training process relies solely on pre-processed data and incurs no real-time computation. As a result, data preparation becomes a one-time cost, since the distance matrix and sample features can be pre-computed and cached, making the computational overhead acceptable and competitive compared to other model-based data selection methods.

\subsection{Case study: neuron-Jaccard retrieval on \textsc{Chem} and \textsc{Phys.}}
\label{app:jaccard_case_study}

To make the reasoning-diversity claim concrete, we run the same $K{=}10$ neuron-Jaccard retriever on two different queries within each of \textsc{Chem} and \textsc{Phys.}, and inspect the retrieved few-shot contexts. The contrast is not in the \emph{overlap} between the two queries' demo sets, which is near-zero given the large sample pool, but in the \emph{internal coherence} of each query's retrieved set.

\paragraph{\textsc{Chem}: retrieval locks onto query-specific reasoning modes.}
Query A asks about iodine-131 half-life calculation. Its retrieved set is dominated by radioactivity, isotope decay, molarity, mass spectrometry, and other per-sample formula-application chemistry tasks. Query B asks about catalytic oxidation depolymerisation. Its retrieved set contains ten demonstrations all concerning polymer breakdown: polymer recycling, controlled-radical-polymerisation reverse processes, dynamic-covalent gel degradation, silicone degradation, and lignin depolymerisation, among others. Each query's retrieved context is internally thematic and distinct from the other query's.

\paragraph{\textsc{Phys.}: retrieval returns generic grab-bags regardless of query.}
Query A asks for the acceleration of a $15\,\mathrm{kg}$ object under a $10\,\mathrm{N}$ force. Its retrieved set is thematically unrelated: an ML-framework question, a stress-intensity-factor method, a heat-engine efficiency problem, a Material-Point-Method description, a specific-heat calculation, a quantum-heat-engine note, a diffuse-interface theory passage, and a PEDOT:PSS conductivity question. Query B asks about the Good Regulator Theorem. Its retrieved set is similarly scattered, spanning van der Waals force, ionizing radiation, cosmic-ray acceleration, resonance-frequency design, CMB temperature, magnetic materials, plasma electron density, and IR polarization. The two sets differ in specific questions but are structurally indistinguishable as generic physics expository or MCQ snippets.

\paragraph{Interpretation.}
In \textsc{Chem}, samples traverse genuinely different reasoning paths, so neuron-level similarity can surface query-specific clusters. In \textsc{Phys.}, samples reuse largely the same reasoning neurons across distinct topics, so $K$-nearest neurons cannot differentiate queries and the retriever collapses toward generic contexts.

\subsection{Relational Analysis of Neuron for Hallucination Detection}

\begin{figure}[h!]
    \centering
    \includegraphics[width=\linewidth]{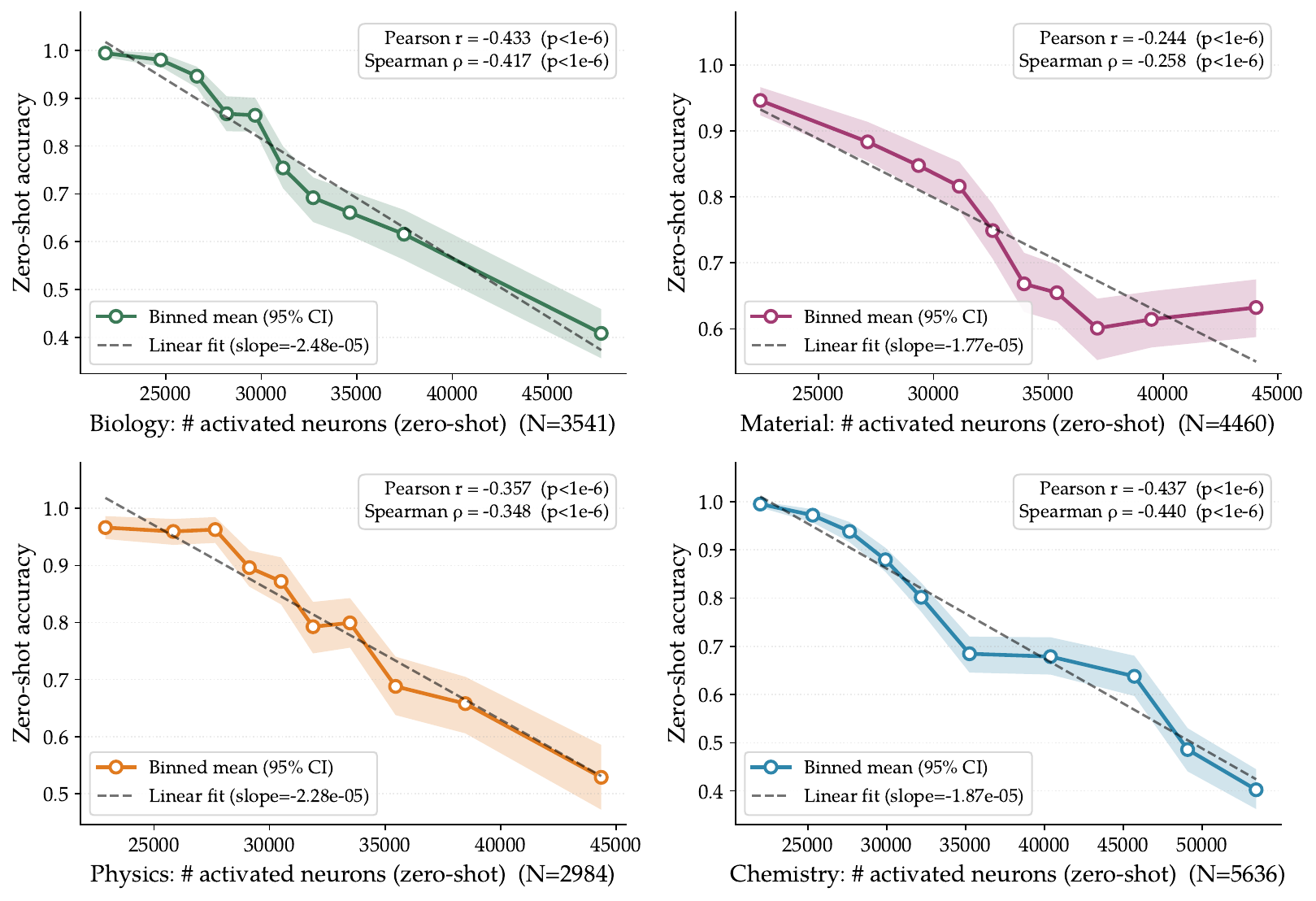}
    \caption{Relational regression results on SciKnowEval.}
    \label{fig:domain-neu-hall-rel}
\end{figure}

\subsection{Additional Results}

\definecolor{posgreen}{rgb}{0.0, 0.55, 0.0}
\definecolor{negred}{rgb}{0.75, 0.0, 0.0}
\begin{table*}[h!]
\centering
\small
\setlength{\tabcolsep}{8pt}
\renewcommand{\arraystretch}{1.15}
\newcolumntype{C}{>{\centering\arraybackslash}p{12mm}}

\scalebox{0.8}{
\begin{tabular}{l|CCCC|C|C|C|C}
\toprule
\multirow{2}{*}{\textbf{Model}} & \multicolumn{5}{c|}{\textbf{SciKnowEval}} & \multirow{2}{*}{\textbf{{Edu.}}} & \multirow{2}{*}{\makecell{\textbf{MMLU-}\\\textbf{Pro}}} & \multirow{2}{*}{\textbf{Avg.}} \\
 & \textsc{Bio} & \textsc{Mat.} & \textsc{Phys.} & \textsc{Chem} & \textit{Avg.} & & & \\
\midrule
Qwen3-4B & 73.98 & 71.12 & 80.20 & 72.11 & 74.35 & 64.95 & 72.16 & 72.42 \\
\midrule
\multicolumn{9}{l}{{LMSI~\citep{huang-etal-2023-large}}} \\
\quad LMSI$_{\textsc{Bio}}$ & \cellcolor{gray!15}74.59$_{\textcolor{posgreen}{+0.61}}$ & 71.22$_{\textcolor{posgreen}{+0.10}}$ & 78.43$_{\textcolor{negred}{-1.77}}$ & 68.60$_{\textcolor{negred}{-3.51}}$ & 73.21$_{\textcolor{negred}{-1.14}}$ & 66.00$_{\textcolor{posgreen}{+1.05}}$ & 58.59$_{\textcolor{negred}{-13.57}}$ & 69.57$_{\textcolor{negred}{-2.85}}$ \\
\quad LMSI$_{\textsc{Mat.}}$ & 72.75$_{\textcolor{negred}{-1.23}}$ & \cellcolor{gray!15}69.08$_{\textcolor{negred}{-2.04}}$ & 73.80$_{\textcolor{negred}{-6.40}}$ & 69.32$_{\textcolor{negred}{-2.79}}$ & 71.23$_{\textcolor{negred}{-3.12}}$ & 60.32$_{\textcolor{negred}{-4.63}}$ & 51.19$_{\textcolor{negred}{-20.97}}$ & 66.08$_{\textcolor{negred}{-6.34}}$ \\
\quad LMSI$_{\textsc{Phys.}}$ & 73.70$_{\textcolor{negred}{-0.28}}$ & 72.30$_{\textcolor{posgreen}{+1.18}}$ & \cellcolor{gray!15}78.54$_{\textcolor{negred}{-1.66}}$ & 71.61$_{\textcolor{negred}{-0.50}}$ & 74.04$_{\textcolor{negred}{-0.31}}$ & 59.49$_{\textcolor{negred}{-5.46}}$ & 58.01$_{\textcolor{negred}{-14.15}}$ & 68.94$_{\textcolor{negred}{-3.48}}$ \\
\quad LMSI$_{\textsc{Chem}}$ & 74.56$_{\textcolor{posgreen}{+0.58}}$ & 72.20$_{\textcolor{posgreen}{+1.08}}$ & 81.40$_{\textcolor{posgreen}{+1.20}}$ & \cellcolor{gray!15}73.72$_{\textcolor{posgreen}{+1.61}}$ & 75.47$_{\textcolor{posgreen}{+1.12}}$ & 61.79$_{\textcolor{negred}{-3.16}}$ & 65.99$_{\textcolor{negred}{-6.17}}$ & 71.61$_{\textcolor{negred}{-0.81}}$ \\
\quad LMSI$_{\textsc{Edu.}}$ & 62.85$_{\textcolor{negred}{-11.13}}$ & 58.44$_{\textcolor{negred}{-12.68}}$ & 64.98$_{\textcolor{negred}{-15.22}}$ & 55.06$_{\textcolor{negred}{-17.05}}$ & 60.34$_{\textcolor{negred}{-14.02}}$ & \cellcolor{gray!15}66.95$_{\textcolor{posgreen}{+2.00}}$ & 49.37$_{\textcolor{negred}{-22.79}}$ & 59.61$_{\textcolor{negred}{-12.81}}$ \\
\quad LMSI$_{\textsc{MMLU}}$ & 74.27$_{\textcolor{posgreen}{+0.29}}$ & 72.63$_{\textcolor{posgreen}{+1.51}}$ & 81.06$_{\textcolor{posgreen}{+0.86}}$ & 71.51$_{\textcolor{negred}{-0.60}}$ & 74.87$_{\textcolor{posgreen}{+0.52}}$ & 67.84$_{\textcolor{posgreen}{+2.89}}$ & \cellcolor{gray!15}68.60$_{\textcolor{negred}{-3.56}}$ & 72.65$_{\textcolor{posgreen}{+0.23}}$ \\
\rowcolor{gray!8}
\quad LMSI$_{\textrm{Avg.}}$ & 72.12$_{\textcolor{negred}{-1.86}}$ & 69.31$_{\textcolor{negred}{-1.81}}$ & 76.37$_{\textcolor{negred}{-3.83}}$ & 68.30$_{\textcolor{negred}{-3.81}}$ & 71.53$_{\textcolor{negred}{-2.83}}$ & 63.73$_{\textcolor{negred}{-1.22}}$ & 58.62$_{\textcolor{negred}{-13.53}}$ & 68.08$_{\textcolor{negred}{-4.34}}$ \\
\midrule

\multicolumn{9}{l}{{TTRL~\citep{zuo_ttrl_2025}}} \\
\quad TTRL$_{\textsc{Bio}}$ & \cellcolor{gray!15}75.00$_{\textcolor{posgreen}{+1.02}}$ & 73.03$_{\textcolor{posgreen}{+1.92}}$ & 83.28$_{\textcolor{posgreen}{+3.09}}$ & 72.85$_{\textcolor{posgreen}{+0.74}}$ & 76.04$_{\textcolor{posgreen}{+1.69}}$ & 65.31$_{\textcolor{posgreen}{+0.36}}$ & 72.08$_{\textcolor{negred}{-0.08}}$ & 73.59$_{\textcolor{posgreen}{+1.17}}$ \\
\quad TTRL$_{\textsc{Mat.}}$ & 74.21$_{\textcolor{posgreen}{+0.22}}$ & \cellcolor{gray!15}73.26$_{\textcolor{posgreen}{+2.14}}$ & 83.25$_{\textcolor{posgreen}{+3.05}}$ & 72.03$_{\textcolor{negred}{-0.08}}$ & 75.69$_{\textcolor{posgreen}{+1.33}}$ & 64.92$_{\textcolor{negred}{-0.03}}$ & 72.20$_{\textcolor{posgreen}{+0.04}}$ & 73.31$_{\textcolor{posgreen}{+0.89}}$ \\
\quad TTRL$_{\textsc{Phys.}}$ & 74.14$_{\textcolor{posgreen}{+0.16}}$ & 72.61$_{\textcolor{posgreen}{+1.49}}$ & \cellcolor{gray!15}82.98$_{\textcolor{posgreen}{+2.79}}$ & 72.61$_{\textcolor{posgreen}{+0.50}}$ & 75.58$_{\textcolor{posgreen}{+1.23}}$ & 65.09$_{\textcolor{posgreen}{+0.14}}$ & 71.94$_{\textcolor{negred}{-0.22}}$ & 73.23$_{\textcolor{posgreen}{+0.81}}$ \\
\quad TTRL$_{\textsc{Chem}}$ & 74.94$_{\textcolor{posgreen}{+0.95}}$ & 72.81$_{\textcolor{posgreen}{+1.69}}$ & 83.40$_{\textcolor{posgreen}{+3.20}}$ & \cellcolor{gray!15}72.43$_{\textcolor{posgreen}{+0.32}}$ & 75.89$_{\textcolor{posgreen}{+1.54}}$ & 64.21$_{\textcolor{negred}{-0.74}}$ & 71.92$_{\textcolor{negred}{-0.24}}$ & 73.28$_{\textcolor{posgreen}{+0.86}}$ \\
\quad TTRL$_{\textsc{Edu.}}$ & 74.56$_{\textcolor{posgreen}{+0.57}}$ & 72.88$_{\textcolor{posgreen}{+1.76}}$ & 83.55$_{\textcolor{posgreen}{+3.35}}$ & 71.91$_{\textcolor{negred}{-0.20}}$ & 75.72$_{\textcolor{posgreen}{+1.37}}$ & \cellcolor{gray!15}64.95$_{\textcolor{posgreen}{+0.00}}$ & 69.43$_{\textcolor{negred}{-2.73}}$ & 72.88$_{\textcolor{posgreen}{+0.46}}$ \\
\quad TTRL$_{\textsc{MMLU}}$ & 74.94$_{\textcolor{posgreen}{+0.95}}$ & 72.88$_{\textcolor{posgreen}{+1.76}}$ & 83.89$_{\textcolor{posgreen}{+3.69}}$ & 72.17$_{\textcolor{posgreen}{+0.06}}$ & 75.97$_{\textcolor{posgreen}{+1.62}}$ & 64.91$_{\textcolor{negred}{-0.04}}$ & \cellcolor{gray!15}72.04$_{\textcolor{negred}{-0.12}}$ & 73.47$_{\textcolor{posgreen}{+1.05}}$ \\
\rowcolor{gray!8}
\quad TTRL$_{\textrm{Avg.}}$ & 74.63$_{\textcolor{posgreen}{+0.65}}$ & 72.91$_{\textcolor{posgreen}{+1.79}}$ & 83.39$_{\textcolor{posgreen}{+3.19}}$ & 72.33$_{\textcolor{posgreen}{+0.22}}$ & 75.82$_{\textcolor{posgreen}{+1.46}}$ & 64.90$_{\textcolor{negred}{-0.05}}$ & 71.60$_{\textcolor{negred}{-0.56}}$ & 73.29$_{\textcolor{posgreen}{+0.87}}$ \\
\midrule

\multicolumn{9}{l}{{Intuitor~\citep{zhao_learning_2025}\footnote{Model trained on the Edu. dataset is dropped on this baseline, since the validation accuracy never goes up.}}} \\
\quad Intuitor$_{\textsc{Bio}}$ & \cellcolor{gray!15}76.21$_{\textcolor{posgreen}{+2.22}}$ & 72.10$_{\textcolor{posgreen}{+0.98}}$ & 82.23$_{\textcolor{posgreen}{+2.03}}$ & 71.31$_{\textcolor{negred}{-0.80}}$ & 75.46$_{\textcolor{posgreen}{+1.11}}$ & 55.70$_{\textcolor{negred}{-9.25}}$ & 66.17$_{\textcolor{negred}{-5.99}}$ & 70.62$_{\textcolor{negred}{-1.80}}$ \\
\quad Intuitor$_{\textsc{Mat.}}$ & 73.98$_{\textcolor{posgreen}{+0.00}}$ & \cellcolor{gray!15}73.41$_{\textcolor{posgreen}{+2.29}}$ & 82.83$_{\textcolor{posgreen}{+2.64}}$ & 71.67$_{\textcolor{negred}{-0.44}}$ & 75.47$_{\textcolor{posgreen}{+1.12}}$ & 63.82$_{\textcolor{negred}{-1.13}}$ & 71.63$_{\textcolor{negred}{-0.53}}$ & 72.89$_{\textcolor{posgreen}{+0.47}}$ \\
\quad Intuitor$_{\textsc{Phys.}}$ & 74.49$_{\textcolor{posgreen}{+0.51}}$ & 72.53$_{\textcolor{posgreen}{+1.41}}$ & \cellcolor{gray!15}83.43$_{\textcolor{posgreen}{+3.24}}$ & 72.07$_{\textcolor{negred}{-0.04}}$ & 75.63$_{\textcolor{posgreen}{+1.28}}$ & 54.81$_{\textcolor{negred}{-10.14}}$ & 67.76$_{\textcolor{negred}{-4.41}}$ & 70.85$_{\textcolor{negred}{-1.57}}$ \\
\quad Intuitor$_{\textsc{Chem}}$ & 74.37$_{\textcolor{posgreen}{+0.38}}$ & 72.20$_{\textcolor{posgreen}{+1.08}}$ & 83.21$_{\textcolor{posgreen}{+3.01}}$ & \cellcolor{gray!15}72.63$_{\textcolor{posgreen}{+0.52}}$ & 75.60$_{\textcolor{posgreen}{+1.25}}$ & 54.94$_{\textcolor{negred}{-10.01}}$ & 69.10$_{\textcolor{negred}{-3.06}}$ & 71.08$_{\textcolor{negred}{-1.34}}$ \\
\quad Intuitor$_{\textsc{MMLU}}$ & 74.11$_{\textcolor{posgreen}{+0.13}}$ & 73.39$_{\textcolor{posgreen}{+2.27}}$ & 83.51$_{\textcolor{posgreen}{+3.31}}$ & 72.57$_{\textcolor{posgreen}{+0.46}}$ & 75.89$_{\textcolor{posgreen}{+1.54}}$ & 62.92$_{\textcolor{negred}{-2.03}}$ & \cellcolor{gray!15}71.91$_{\textcolor{negred}{-0.25}}$ & 73.07$_{\textcolor{posgreen}{+0.65}}$ \\
\rowcolor{gray!8}
\quad Intuitor$_{\textrm{Avg.}}$ & 74.63$_{\textcolor{posgreen}{+0.65}}$ & 72.73$_{\textcolor{posgreen}{+1.61}}$ & 83.04$_{\textcolor{posgreen}{+2.85}}$ & 72.05$_{\textcolor{negred}{-0.06}}$ & 75.61$_{\textcolor{posgreen}{+1.26}}$ & 58.44$_{\textcolor{negred}{-6.51}}$ & 69.32$_{\textcolor{negred}{-2.85}}$ & 71.70$_{\textcolor{negred}{-0.72}}$ \\
\midrule
\multicolumn{9}{l}{{Neuron-OPSD (ours)}} \\
\quad N-OPSD$_{\textsc{Bio}}$ & \cellcolor{gray!15}74.02$_{\textcolor{posgreen}{+0.03}}$ & 72.96$_{\textcolor{posgreen}{+1.84}}$ & 82.87$_{\textcolor{posgreen}{+2.67}}$ & 71.97$_{\textcolor{negred}{-0.14}}$ & 75.45$_{\textcolor{posgreen}{+1.10}}$ & 66.31$_{\textcolor{posgreen}{+1.36}}$ & 69.46$_{\textcolor{negred}{-2.71}}$ & 72.93$_{\textcolor{posgreen}{+0.51}}$ \\
\quad N-OPSD$_{\textsc{Mat.}}$ & 73.83$_{\textcolor{negred}{-0.16}}$ & \cellcolor{gray!15}73.39$_{\textcolor{posgreen}{+2.27}}$ & 83.13$_{\textcolor{posgreen}{+2.94}}$ & 72.83$_{\textcolor{posgreen}{+0.72}}$ & 75.79$_{\textcolor{posgreen}{+1.44}}$ & 65.34$_{\textcolor{posgreen}{+0.39}}$ & 71.22$_{\textcolor{negred}{-0.94}}$ & 73.29$_{\textcolor{posgreen}{+0.87}}$ \\
\quad N-OPSD$_{\textsc{Phys.}}$ & 74.78$_{\textcolor{posgreen}{+0.79}}$ & 73.71$_{\textcolor{posgreen}{+2.60}}$ & \cellcolor{gray!15}83.13$_{\textcolor{posgreen}{+2.94}}$ & 71.93$_{\textcolor{negred}{-0.18}}$ & 75.89$_{\textcolor{posgreen}{+1.54}}$ & 65.20$_{\textcolor{posgreen}{+0.25}}$ & 71.24$_{\textcolor{negred}{-0.92}}$ & 73.33$_{\textcolor{posgreen}{+0.91}}$ \\
\quad N-OPSD$_{\textsc{Chem}}$ & 74.24$_{\textcolor{posgreen}{+0.26}}$ & 73.69$_{\textcolor{posgreen}{+2.57}}$ & 83.21$_{\textcolor{posgreen}{+3.01}}$ & \cellcolor{gray!15}72.03$_{\textcolor{negred}{-0.08}}$ & 75.79$_{\textcolor{posgreen}{+1.44}}$ & 65.45$_{\textcolor{posgreen}{+0.50}}$ & 71.36$_{\textcolor{negred}{-0.80}}$ & 73.33$_{\textcolor{posgreen}{+0.91}}$ \\
\quad N-OPSD$_{\textsc{Edu.}}$ & 75.13$_{\textcolor{posgreen}{+1.14}}$ & 73.14$_{\textcolor{posgreen}{+2.02}}$ & 83.02$_{\textcolor{posgreen}{+2.82}}$ & 72.85$_{\textcolor{posgreen}{+0.74}}$ & 76.03$_{\textcolor{posgreen}{+1.68}}$ & \cellcolor{gray!15}72.19$_{\textcolor{posgreen}{+7.24}}$ & 71.96$_{\textcolor{negred}{-0.21}}$ & 74.71$_{\textcolor{posgreen}{+2.29}}$ \\
\quad N-OPSD$_{\textsc{MMLU}}$ & 74.43$_{\textcolor{posgreen}{+0.44}}$ & 73.46$_{\textcolor{posgreen}{+2.34}}$ & 83.47$_{\textcolor{posgreen}{+3.28}}$ & 73.03$_{\textcolor{posgreen}{+0.92}}$ & 76.10$_{\textcolor{posgreen}{+1.75}}$ & 64.33$_{\textcolor{negred}{-0.62}}$ & \cellcolor{gray!15}72.04$_{\textcolor{negred}{-0.12}}$ & 73.46$_{\textcolor{posgreen}{+1.04}}$ \\
\rowcolor{gray!8}
\quad N-OPSD$_{\textrm{Avg.}}$ & 74.40$_{\textcolor{posgreen}{+0.42}}$ & 73.39$_{\textcolor{posgreen}{+2.27}}$ & 83.14$_{\textcolor{posgreen}{+2.94}}$ & 72.44$_{\textcolor{posgreen}{+0.33}}$ & 75.84$_{\textcolor{posgreen}{+1.49}}$ & 66.47$_{\textcolor{posgreen}{+1.52}}$ & 71.21$_{\textcolor{negred}{-0.95}}$ & 73.51$_{\textcolor{posgreen}{+1.09}}$ \\
\bottomrule
\end{tabular}
}

\caption{
Cross-domain evaluation on Qwen3-4B-Thinking-2507, performance reported in Avg@8. Each row reports a single model trained on one source domain and evaluated on all remaining test sets. Subscripts give the $\Delta$ vs.\ the {untrained} baseline.
}
\label{tab:cross_domain_main_avg8}
\end{table*}

\definecolor{posgreen}{rgb}{0.0, 0.55, 0.0}
\definecolor{negred}{rgb}{0.75, 0.0, 0.0}
\begin{table*}[t]
\centering
\small
\setlength{\tabcolsep}{6pt}
\renewcommand{\arraystretch}{1.15}
\newcolumntype{C}{>{\centering\arraybackslash}p{13mm}}
\scalebox{0.85}{
\begin{tabular}{l|CCCC|C|C|C|C}
\toprule
\multirow{2}{*}{\textbf{Model}} & \multicolumn{5}{c|}{\textbf{SciKnowEval}} & \multirow{2}{*}{\textbf{\textsc{Edu.}}} & \multirow{2}{*}{\makecell{\textbf{MMLU-}\\\textbf{Pro}}} & \multirow{2}{*}{\textbf{Avg.}} \\
 & \textsc{Bio} & \textsc{Mat.} & \textsc{Phys.} & \textsc{Chem} & \textit{Avg.} & & & \\
\midrule
Qwen3-4B-think & 0.195 & 0.213 & 0.095 & 0.173 & 0.169 & 0.246 & 0.184 & 0.184 \\
\midrule
\multicolumn{9}{l}{{TTRL~\citep{zuo_ttrl_2025}}} \\
\quad TTRL$_{\textsc{Bio}}$ & \cellcolor{gray!15}0.204$_{\textcolor{negred}{+0.009}}$ & 0.220$_{\textcolor{negred}{+0.007}}$ & 0.118$_{\textcolor{negred}{+0.023}}$ & 0.169$_{\textcolor{posgreen}{-0.004}}$ & 0.178$_{\textcolor{negred}{+0.009}}$ & 0.247$_{\textcolor{negred}{+0.000}}$ & 0.178$_{\textcolor{posgreen}{-0.006}}$ & 0.189$_{\textcolor{negred}{+0.005}}$ \\
\quad TTRL$_{\textsc{Mat.}}$ & 0.203$_{\textcolor{negred}{+0.008}}$ & \cellcolor{gray!15}0.208$_{\textcolor{posgreen}{-0.005}}$ & 0.124$_{\textcolor{negred}{+0.029}}$ & 0.181$_{\textcolor{negred}{+0.008}}$ & 0.179$_{\textcolor{negred}{+0.010}}$ & 0.242$_{\textcolor{posgreen}{-0.005}}$ & 0.186$_{\textcolor{negred}{+0.002}}$ & 0.191$_{\textcolor{negred}{+0.006}}$ \\
\quad TTRL$_{\textsc{Phys.}}$ & 0.214$_{\textcolor{negred}{+0.018}}$ & 0.215$_{\textcolor{negred}{+0.002}}$ & \cellcolor{gray!15}0.123$_{\textcolor{negred}{+0.028}}$ & 0.166$_{\textcolor{posgreen}{-0.006}}$ & 0.180$_{\textcolor{negred}{+0.011}}$ & 0.239$_{\textcolor{posgreen}{-0.008}}$ & 0.188$_{\textcolor{negred}{+0.004}}$ & 0.191$_{\textcolor{negred}{+0.007}}$ \\
\quad TTRL$_{\textsc{Chem}}$ & 0.204$_{\textcolor{negred}{+0.009}}$ & 0.223$_{\textcolor{negred}{+0.010}}$ & 0.118$_{\textcolor{negred}{+0.023}}$ & \cellcolor{gray!15}0.192$_{\textcolor{negred}{+0.019}}$ & 0.184$_{\textcolor{negred}{+0.015}}$ & 0.251$_{\textcolor{negred}{+0.005}}$ & 0.197$_{\textcolor{negred}{+0.013}}$ & 0.198$_{\textcolor{negred}{+0.013}}$ \\
\quad TTRL$_{\textsc{Edu.}}$ & 0.199$_{\textcolor{negred}{+0.003}}$ & 0.215$_{\textcolor{negred}{+0.003}}$ & 0.111$_{\textcolor{negred}{+0.015}}$ & 0.164$_{\textcolor{posgreen}{-0.009}}$ & 0.172$_{\textcolor{negred}{+0.003}}$ & \cellcolor{gray!15}0.248$_{\textcolor{negred}{+0.002}}$ & 0.202$_{\textcolor{negred}{+0.018}}$ & 0.190$_{\textcolor{negred}{+0.005}}$ \\
\quad TTRL$_{\textsc{MMLU}}$ & 0.200$_{\textcolor{negred}{+0.005}}$ & 0.224$_{\textcolor{negred}{+0.011}}$ & 0.114$_{\textcolor{negred}{+0.019}}$ & 0.169$_{\textcolor{posgreen}{-0.003}}$ & 0.177$_{\textcolor{negred}{+0.008}}$ & 0.243$_{\textcolor{posgreen}{-0.004}}$ & \cellcolor{gray!15}0.198$_{\textcolor{negred}{+0.014}}$ & 0.191$_{\textcolor{negred}{+0.007}}$ \\
\rowcolor{gray!8}
\quad TTRL$_{\textrm{Avg.}}$ & 0.204$_{\textcolor{negred}{+0.009}}$ & 0.218$_{\textcolor{negred}{+0.005}}$ & 0.118$_{\textcolor{negred}{+0.023}}$ & 0.174$_{\textcolor{negred}{+0.001}}$ & 0.178$_{\textcolor{negred}{+0.009}}$ & 0.245$_{\textcolor{posgreen}{-0.002}}$ & 0.191$_{\textcolor{negred}{+0.007}}$ & 0.192$_{\textcolor{negred}{+0.007}}$ \\
\midrule
\multicolumn{9}{l}{{Intuitor~\citep{zhao_learning_2025}}} \\
\quad Intuitor$_{\textsc{Bio}}$ & \cellcolor{gray!15}0.208$_{\textcolor{negred}{+0.013}}$ & 0.236$_{\textcolor{negred}{+0.023}}$ & 0.141$_{\textcolor{negred}{+0.046}}$ & 0.210$_{\textcolor{negred}{+0.038}}$ & 0.199$_{\textcolor{negred}{+0.030}}$ & 0.312$_{\textcolor{negred}{+0.066}}$ & 0.229$_{\textcolor{negred}{+0.045}}$ & 0.223$_{\textcolor{negred}{+0.039}}$ \\
\quad Intuitor$_{\textsc{Mat.}}$ & 0.209$_{\textcolor{negred}{+0.014}}$ & \cellcolor{gray!15}0.212$_{\textcolor{posgreen}{-0.001}}$ & 0.124$_{\textcolor{negred}{+0.029}}$ & 0.166$_{\textcolor{posgreen}{-0.007}}$ & 0.178$_{\textcolor{negred}{+0.009}}$ & 0.246$_{\textcolor{negred}{+0.000}}$ & 0.188$_{\textcolor{negred}{+0.004}}$ & 0.191$_{\textcolor{negred}{+0.007}}$ \\
\quad Intuitor$_{\textsc{Phys.}}$ & 0.213$_{\textcolor{negred}{+0.018}}$ & 0.229$_{\textcolor{negred}{+0.016}}$ & \cellcolor{gray!15}0.117$_{\textcolor{negred}{+0.021}}$ & 0.176$_{\textcolor{negred}{+0.004}}$ & 0.184$_{\textcolor{negred}{+0.015}}$ & 0.341$_{\textcolor{negred}{+0.095}}$ & 0.210$_{\textcolor{negred}{+0.026}}$ & 0.214$_{\textcolor{negred}{+0.030}}$ \\
\quad Intuitor$_{\textsc{Chem}}$ & 0.213$_{\textcolor{negred}{+0.017}}$ & 0.225$_{\textcolor{negred}{+0.012}}$ & 0.130$_{\textcolor{negred}{+0.034}}$ & \cellcolor{gray!15}0.198$_{\textcolor{negred}{+0.026}}$ & 0.191$_{\textcolor{negred}{+0.022}}$ & 0.319$_{\textcolor{negred}{+0.073}}$ & 0.234$_{\textcolor{negred}{+0.050}}$ & 0.220$_{\textcolor{negred}{+0.036}}$ \\
\quad Intuitor$_{\textsc{MMLU}}$ & 0.197$_{\textcolor{negred}{+0.002}}$ & 0.205$_{\textcolor{posgreen}{-0.008}}$ & 0.113$_{\textcolor{negred}{+0.018}}$ & 0.142$_{\textcolor{posgreen}{-0.031}}$ & 0.164$_{\textcolor{posgreen}{-0.005}}$ & 0.267$_{\textcolor{negred}{+0.021}}$ & \cellcolor{gray!15}0.174$_{\textcolor{posgreen}{-0.010}}$ & 0.183$_{\textcolor{posgreen}{-0.001}}$ \\
\rowcolor{gray!8}
\quad Intuitor$_{\textrm{Avg.}}$ & 0.208$_{\textcolor{negred}{+0.013}}$ & 0.221$_{\textcolor{negred}{+0.008}}$ & 0.125$_{\textcolor{negred}{+0.030}}$ & 0.179$_{\textcolor{negred}{+0.006}}$ & 0.183$_{\textcolor{negred}{+0.014}}$ & 0.297$_{\textcolor{negred}{+0.051}}$ & 0.207$_{\textcolor{negred}{+0.023}}$ & 0.206$_{\textcolor{negred}{+0.022}}$ \\
\midrule
\multicolumn{9}{l}{{Neuron-OPSD, ours}} \\
\quad N-OPSD$_{\textsc{Bio}}$ & \cellcolor{gray!15}0.193$_{\textcolor{posgreen}{-0.003}}$ & 0.199$_{\textcolor{posgreen}{-0.014}}$ & 0.111$_{\textcolor{negred}{+0.015}}$ & 0.159$_{\textcolor{posgreen}{-0.013}}$ & 0.165$_{\textcolor{posgreen}{-0.004}}$ & 0.221$_{\textcolor{posgreen}{-0.025}}$ & 0.194$_{\textcolor{negred}{+0.010}}$ & 0.180$_{\textcolor{posgreen}{-0.005}}$ \\
\quad N-OPSD$_{\textsc{Mat.}}$ & 0.206$_{\textcolor{negred}{+0.011}}$ & \cellcolor{gray!15}0.207$_{\textcolor{posgreen}{-0.006}}$ & 0.121$_{\textcolor{negred}{+0.026}}$ & 0.156$_{\textcolor{posgreen}{-0.017}}$ & 0.173$_{\textcolor{negred}{+0.004}}$ & 0.219$_{\textcolor{posgreen}{-0.028}}$ & 0.185$_{\textcolor{negred}{+0.001}}$ & 0.182$_{\textcolor{posgreen}{-0.002}}$ \\
\quad N-OPSD$_{\textsc{Phys.}}$ & 0.196$_{\textcolor{negred}{+0.001}}$ & 0.198$_{\textcolor{posgreen}{-0.015}}$ & \cellcolor{gray!15}0.114$_{\textcolor{negred}{+0.019}}$ & 0.170$_{\textcolor{posgreen}{-0.003}}$ & 0.170$_{\textcolor{negred}{+0.001}}$ & 0.217$_{\textcolor{posgreen}{-0.029}}$ & 0.189$_{\textcolor{negred}{+0.005}}$ & 0.181$_{\textcolor{posgreen}{-0.004}}$ \\
\quad N-OPSD$_{\textsc{Chem}}$ & 0.198$_{\textcolor{negred}{+0.003}}$ & 0.215$_{\textcolor{negred}{+0.002}}$ & 0.117$_{\textcolor{negred}{+0.022}}$ & \cellcolor{gray!15}0.171$_{\textcolor{negred}{-0.002}}$ & 0.175$_{\textcolor{negred}{+0.006}}$ & 0.225$_{\textcolor{posgreen}{-0.021}}$ & 0.180$_{\textcolor{posgreen}{-0.004}}$ & 0.184$_{\textcolor{negred}{+0.000}}$ \\
\quad N-OPSD$_{\textsc{Edu.}}$ & 0.195$_{\textcolor{posgreen}{-0.000}}$ & 0.218$_{\textcolor{negred}{+0.005}}$ & 0.122$_{\textcolor{negred}{+0.027}}$ & 0.171$_{\textcolor{posgreen}{-0.002}}$ & 0.177$_{\textcolor{negred}{+0.007}}$ & \cellcolor{gray!15}0.191$_{\textcolor{posgreen}{-0.056}}$ & 0.179$_{\textcolor{posgreen}{-0.005}}$ & 0.179$_{\textcolor{posgreen}{-0.005}}$ \\
\quad N-OPSD$_{\textsc{MMLU}}$ & 0.206$_{\textcolor{negred}{+0.010}}$ & 0.215$_{\textcolor{negred}{+0.002}}$ & 0.113$_{\textcolor{negred}{+0.018}}$ & 0.158$_{\textcolor{posgreen}{-0.014}}$ & 0.173$_{\textcolor{negred}{+0.004}}$ & 0.239$_{\textcolor{posgreen}{-0.008}}$ & \cellcolor{gray!15}0.180$_{\textcolor{posgreen}{-0.004}}$ & 0.185$_{\textcolor{negred}{+0.001}}$ \\
\rowcolor{gray!8}
\quad N-OPSD$_{\textrm{Avg.}}$ & 0.199$_{\textcolor{negred}{+0.004}}$ & 0.209$_{\textcolor{posgreen}{-0.004}}$ & 0.116$_{\textcolor{negred}{+0.021}}$ & 0.164$_{\textcolor{posgreen}{-0.009}}$ & 0.172$_{\textcolor{negred}{+0.003}}$ & 0.219$_{\textcolor{posgreen}{-0.027}}$ & 0.185$_{\textcolor{negred}{+0.001}}$ & 0.182$_{\textcolor{posgreen}{-0.002}}$ \\
\bottomrule
\end{tabular}
}
\caption{Expected Calibration Error (ECE), where lower is better, on the same test sets as Tab.~\ref{tab:cross_domain_main_avg8}. Each cell shows the absolute ECE with the delta vs the \textit{Base} row in subscript, with \textcolor{posgreen}{green} for lower ECE and better calibration and \textcolor{negred}{red} for worse calibration. Gray cells mark the source domain. Shaded rows show the column-wise average across the trained models per method.}
\label{tab:cross_domain_ece_avg8}
\end{table*}

\end{document}